%% file: DeepSpecularIntrinsics.tex
\documentclass[runningheads]{llncs}
\usepackage{graphicx}
\usepackage{amsmath,amssymb}
\usepackage{color}
\usepackage{float}
\usepackage{booktabs}
\usepackage{microtype}
\usepackage[table]{xcolor}
\usepackage[width=122mm,left=12mm,paperwidth=146mm,height=193mm,top=12mm,paperheight=217mm]{geometry}
\usepackage{wrapfig}
\usepackage{authblk}
\usepackage{hyperref}

\begin{document}
\pagestyle{headings}
\mainmatter
\def\ECCV16SubNumber{712}

\include{our-commands}

\DeclareGraphicsExtensions{.pdf,.ai,.eps}
\DeclareGraphicsRule{.ai}{pdf}{.ai}{}
\DeclareGraphicsRule{.eps}{pdf}{.eps}{}

\title{DeLight-Net: Decomposing Reflectance Maps into Specular Materials and Natural Illumination}

\titlerunning{Delight-Net: Decomposing Reflectance Maps into Materials and Illumination}

\authorrunning{Georgoulis, Rematas, Ritschel, Fritz, Van Gool, Tuytelaars}

\author{Stamatios Georgoulis$^{*1}$ \and Konstantinos Rematas$^{*1}$ \and Tobias Ritschel$^{2}$ \and \\ Mario Fritz$^{3}$ \and Luc Van Gool$^{1}$ \and Tinne Tuytelaars$^{1}$}
\institute{$^{1}$KU Leuven\qquad $^{2}$University College London \qquad $^{3}$MPI Informatics}

\maketitle
\begin{abstract} 

In this paper we are extracting surface reflectance and natural environmental illumination from a reflectance map, \ie from a single 2D image of a sphere of one material under one illumination.
This is a notoriously difficult problem, yet key to various re-rendering applications.
With the recent advances in estimating reflectance maps from 2D images their further decomposition has become increasingly relevant.

To this end, we propose a Convolutional Neural Network (CNN) architecture to reconstruct both material parameters (\ie Phong) as well as illumination (\ie high-resolution spherical illumination maps), that is solely trained on synthetic data.
We demonstrate that decomposition of synthetic as well as real photographs of reflectance maps, both in High Dynamic Range (HDR), and, for the first time, on Low Dynamic Range (LDR) as well.
Results are compared to previous approaches  quantitatively as well as qualitatively in terms of re-renderings where illumination, material, view or shape are changed.

\end{abstract}

\mysection{Introduction}{Introduction}
Re-rendering the scene in an image after changing view, object poses or illumination, requires estimating its {\em intrinsic properties}~\cite{Barrow1978}.\blfootnote{$^{*}$ Denotes equal contribution.}
These are the physical quantities yielding a scene's appearance through their interaction: the incoming illumination, the reflectance function (\ie material), and the 3D shape (\ie normals).
With the intrinsics at our disposal, we can then freely replace or alter one of them and re-render the scene (see \refFig{Concept}). 

\mycfigure{Concept}{
Overview of our approach in an application context \emph{(From left to right)}.
Starting from an image, a reflectance map is generated using an off-the-shelf method \emph{(black arrow)}.
Next, it is split into illumination and reflectance \emph{(pink arrow)} by means of two CNNs.
These components can then be manipulated independently \emph{(blue arrow)}, such as exchanged, or shape and view could be altered \emph{(not shown)}.
The final result is achieved by re-synthesizing a new reflectance map and from this, a new image \emph{(yellow arrow)}.
}

Extracting the intrinsic properties is, unfortunately, a difficult problem as e.g. many combinations of illumination and reflectance will result in the same appearance (see \refSec{PreviousWork} for proposed solutions). 
Here, we follow the approach where, in a first step, the 3D geometry and reflectance maps are estimated, and then, in a second step, these reflectance maps are decomposed further into reflectance and illumination. 
For the first step, a number of solutions have been proposed: they can either be estimated via intermediate depth \cite{Eigen2014a,Li2015,Liu2015} as well as normals \cite{Eigen2015,Wang2015,Li2015,Richter2015} or even acquired directly \cite{Rematas2016}.
In other cases, spherical material probes are given as photographs, or sometimes manually painted by artists \cite{Zbrush}.

In this paper we focus on the second step, \ie extracting, from the reflectance map, the illumination and reflectance. In particular, we propose to use a Convolutional Neural Network (CNN) to carry out the above decomposition, as they have shown unprecedented performance in de-convolution tasks with similar requirements.
Moreover, being a data-driven approach, they can easily incorporate priors about illumination and reflectance statistics.
We make the simplifying assumption that the 2D image we start from shows a reflectance map, \ie the orthographic projection of a sphere covered with a single material (same surface reflectance properties everywhere), illuminated as in a natural, complex scene. 

\mysection{Previous Work}{PreviousWork}
\emph{Intrinsics} are the individual physical properties that yield a scene's appearance through their interaction~\cite{Barrow1978}.
As an example, incoming illumination reflected in the direction of the observer yields an appearance influenced by both.
3D~shape is another intrinsic property of the objects in a scene, that also influences appearance through the surface's orientation (normals).
Ideally, one can retrieve all these pieces of the appearance jigsaw puzzle separately.
In practice, even if one fixes a single component (shape, reflectance, or illumination) by assuming it known or simple, what one is left with is still a hard decomposition problem for the remaining two.
Sometimes one also keeps two of the three intertwined, only retrieving the third as a separate entity. 

As making assumptions about one or more of the intrinsics is important to get a handle on the decomposition problem, it also is relevant to better understand their natural statistics.
Databases of reflectance \cite{Dana1999} or illumination \cite{Debevec1998,Dror2001} samples have allowed to acquire such statistics, but exploiting them in computation remains challenging.
Recent databases focus on images captured in the wild, e.g. annotated for reflectance using crowd-sourcing \cite{Bell2014}.

Next, we describe related work, that we mainly found in the three main research strands listed below.
The following discussion gradually homes in on work that gets closer and closer to ours. 

\paragraph{Factoring Intrinsics out of Images.}

\emph{Shape-from-shading} is probably one of the oldest attempts to get intrinsic information from single images \cite{Barrow1978}.
The goal is to extract the shape intrinsic from appearance, eventually leading to an orientation map or even a full 3D surface.
Typically, strong assumptions on both the surface reflectance and the illumination were made.

Recent, seminal work by Baron and Malik~\cite{Barron2015a} decompose images into shape, reflectance and illumination, but only for scalar reflectance (i.e. diffuse albedo) and lower illumination frequencies.
Assuming none of the 3 components known was a great feat nonetheless.
Assuming one component known, allowed Lombardi and Nishino~\cite{Lombardi2015} as well as Johnson and Adelson~\cite{Johnson2011} to move beyond diffuse albedo, and also consider natural illumination.
Interestingly, they found that natural illumination can simplify rather than complicate the remaining decomposition.

In this work we address a problem similar to Lombardi and Nishino~\cite{Lombardi2015}: a sphere with a single, unknown material on the surface (homogeneous surface reflectance) is observed under some natural illumination.
Hence, the shape is known, and the reflectance and illumination remain to be separately retrieved. Addressing the same problem as previous work, our solution is fundamentally different: instead of seeking to invert the physical process under the guidance of a manually designed - thus limiting - prior, our work entirely relies on data to learn the backward mapping from a reflectance map to its intrinsics.
Our results indicate this inverse mapping can be learned, leading to high-quality, detailed, yet naturalistic illumination maps. The underlying network has learned tricks such that windows are bright or that it is the sky that is blue and not so much the object. 
Moreover, our approach is the first to perform a slightly altered task, that is much closer to practice, where the image to decompose is captured using a Low Dynamic Range (LDR) sensor, yet the resulting illumination map is High Dynamic Range (HDR) as required for re-synthesis, even if the synthetic outcome is LDR again.
Previous work typically has considered either HDR input or produced only LDR illumination maps.

\paragraph{Reflectance maps.}
It is not always required to separate reflectance and illumination.
\emph{Reflectance maps} \cite{Horn1979} - that assign an appearance (i.e. RGB color) to a surface orientation, thereby combining reflectance and illumination - suffice for many important applications.
Examples are novel view generation (if the 3D shape is available) or material exchanges.
Such reflectance maps can be obtained in multiple ways, e.g. using Internet photo collections of diffuse objects to produce a rough 3D shape and then extracting reflectance maps in a second step \cite{Haber2009}.
A recent approach by Richter and Roth~\cite{Richter2015} first estimates a diffuse reflectance map using approximate normals and then refines the normal map using the reflectance map as a guide.
Different from our approach, they assume diffuse surfaces to be approximated using 2nd-order spherical harmonics (SH) and learn to refine the normals from the reflectance map using a regression forest.
Rematas~\citeetal{Rematas2016}, have used CNNs to directly compute a reflectance map from an image showing objects of an approximately known shape (known object class, e.g. cars), but without further decomposing it into reflectance and illumination.

In computer graphics, reflectance maps are popular to capture, transfer and manipulate the orientation-dependent appearance of photo-realistic or artistic shading.
They are also known as ``lit spheres'' \cite{Sloan2001} or ``MatCaps'' \cite{Zbrush}.
A special user interface is required to map surface orientation to appearance at sparse points in an image, from which orientations are interpolated for in-between pixels to fill the lit sphere (e.g. \cite{RematasCVPR14} manually aligned a 3D model with an image to generate reflectance maps).
Khan~\citeetal{Khan2006} made small diffuse objects in a single cluttered image to appear specular or transparent, using image manipulations that included manual intervention.
This approach works surprisingly well visually, but no larger material or illumination changes are possible and the results deviate substantially from ground-truth.
Our results do not just look plausible, but stay closer to ground-truth even when scene parameters are changed substantially.
Surface reflectance and illumination are separated in our case.

While reflectance maps are easy to use, the fact that they combine reflectance and illumination limits their use in practice: it is not possible to manipulate illumination alone, nor is it possible to change reflectance.
While simple manipulations like color changes are plausible, other manipulations, such as increasing glossiness - which reveals information when becoming closer to a mirror - are not practical.
Reflectance maps also do not separate view-dependent and view-independent shading.
As a result, diffuse shading might change when the view changes, although it should be view-independent.
Symmetrically, specular highlights that are view-dependent in reality can become view-independent, making them appear to be ``painted'' onto objects.
In summary, the price for such simple acquisition and synthesis based on reflectance maps is that desirable operations like the rotation of illumination or view changes are not possible. They call for a genuine decomposition of the reflectance maps into their intrinsics, as done here.

\paragraph{Deep Learning.}
Our method uses Convolutional Neural Networks (CNNs), hence it is useful to also highlight the relevant literature part. Recent CNNs have shown strong performance on inverse tasks like ours.
Based on ideas of encoding-decoding strategies similar to auto-encoders, convolutional decoders have been developed \cite{zeiler2010deconvolutional,lee2009convolutional} to decode condensed representations back to images.
They are useful when a per-pixel prediction is required. For instance, Long~\citeetal{Long2015} and Hariharana~\citeetal{hypercolumn} used this paradigm for semantic image segmentation.
Dosovitskiy~\etal synthesised new object views, given object class, view and view transformations as input.
Similarly, Kulkarni~\citeetal{Kulkarni2014} propose \emph{deep convolutional inverse graphics networks}, also with an encoder-decoder architecture, that given an image can synthesize novel views.
In contrast, our approach achieves a new mapping not to an image, but to spherical maps representing intrinsic properties - from reflectance map to surface reflectance and environmental illumination.

\emph{Deep lambertian networks}: Tang~\citeetal{Tang2012} apply deep belief networks for the joint estimation of lambertian reflectance, an orientation map and the direction of a point light source.
They rely on Gaussian Restricted Boltzmann Machines to model the prior of albedo and surface normals for inference from a single image.
In contrast, we address specular materials under general illumination.

Another branch of research proposes to use neural networks for
depth estimation~\cite{Eigen2014a,Li2015,Liu2015},
normal estimation~\cite{Eigen2015,Wang2015,Li2015},
intrinsic image decomposition\cite{Narihira2015b,Zhou2015} and
lightness \cite{Narihira2015a}.
Wang~\citeetal{Wang2015} show that a careful mixture of deep architectures with hand-engineered models allow for accurate surface normal estimation.
Observing that normals, depth and segmentations are related tasks, Eigen~\citeetal{Eigen2015} propose a coarse-to-fine, multi-scale and multi-purpose deep network that jointly optimizes depth and normal estimation and semantic segmentation.
Likewise, Li~\citeetal{Li2015} apply deep regression using convolutional neural networks for depth and normal estimation, the output of which is refined further by a conditional random field.
Going one step further, Liu~\citeetal{Liu2015} propose to embed both the unary and the pairwise potentials of a conditional random field in a unified deep network.
In contrast, our goal is not extracting depths, normals or reflectance maps from images, but decomposing the appearance of a known spherical geometry into the underlying surface reflectance and environmental illumination.

\mysection{DeLight-Net}{OurAppraoch}

\mysubsection{Overview}{Overview}

The input to our approach is a reflectance map, see \refFig{Concept}. This image can be HDR or LDR. A reflectance map can be produced in several ways: photos, surface estimation, or using a CNN.
For the purpose of this paper, we assume the reflectance map to be given, but it is useful keeping these options in mind. We therefore further elaborate on them. When a spherical sample of the desired material is available, it can directly be put under the desired illumination and photographed.
In practice, this is usually not the case - the sample has a different shape.
If the shape is known, its normals are known, and its reflectance map can be retrieved, at least for all observed surface orientations.
Several options exist to acquire the sample's shape, including 3D scanning, depth estimation, eventually using a CNN \cite{Eigen2014a,Li2015,Liu2015} or directly estimating the normals \cite{Eigen2015,Wang2015,Li2015}.
In other cases, the reflectance map can directly be estimated from the image using a CNN \cite{Rematas2016}.

The outputs of our method are Phong reflectance parameters and an HDR environment illumination map.
Other parametric reflectance models would be possible as well, but this remains future work.
The illumination map is an HDR spherical image, expressing illumination's directional dependency.
HDR is a critical property to have for illumination \cite{Debevec1996,Dror2001}, as without it re-illumination is likely to fail in many real-world cases.
Note, that our estimated illumination is still HDR even when the input is only LDR.

We enable this mapping by proposing \emph{DeLight-Net} which consists of two parallel CNNs, trained on the same synthetic data.
Both CNNs take as input the reflectance map.
The first (illumination network) outputs the HDR illumination map with its spherical parametrization.
The second (reflectance network) outputs a parameter vector, which is 7-dimensional in the case of the Phong reflection model: one color for diffuse and one for specular, and a glossiness value, defining how shiny the material is.
Furthermore, we propose two variants: the first variant shares intermediate representations to perform the estimation jointly; the second, is a special up-sampling scheme to ensure sharp edges in the illumination.

Next, after giving more formal definitions of the entities used in \refSec{Definitions}, we will describe the training data (\refSec{TrainingData}), and the particular network structure found to be most effective for this task (\refSec{NetworkStructure}).
The section concludes by describing the direct estimation of Phong reflectance from a reflectance map if the illumination is given (\refSec{Direct}).

\mysubsection{Definitions}{Definitions}
Here, we recall the definition of the rendering equation, leading to the notion of reflectance maps, as well as the surface reflectance model that we will use.
The rendering equation \cite{Kajiya1986} (RE) states, that for one wavelength
\[
\tilde
L_\mathrm o(\mathbf x,\omega_\mathrm o)=
\tilde L_\mathrm e(\mathbf x,\omega_\mathrm o)+
\int_{\Omega^+}
\tilde L_\mathrm i(\mathbf x,\omega_\mathrm o)
\tilde f_\mathrm r(\mathbf x,\omega_\mathrm o,\omega_\mathrm i)
\left<
n(\mathbf x),
\omega_\mathrm i
\right>^+
\mathrm d \omega_\mathrm i,
\]
where
$L_\mathrm o$ is the outgoing radiance,
$L_\mathrm e$ the emitted radiance,
$L_\mathrm i$ the incoming radiance,
$f_\mathrm r$ the Bidirectional Reflectance Distribution Function (BRDF) and $n(\mathbf x)$ the surface orientation.
The radiances are both function of position and direction.
The reflected part is the intergal over the upper hemisphere of the product of incoming light, BRDF and the dot product of surface normal and integration direction.
In this work it is assumed that
\emph{i)} there is no emission
\emph{ii)} there is only one material (one surface reflectance model to be considered),
\emph{iii)} the shape is spherical,
\emph{iv)} seen under orthographic projection from an infinitely far-away observer, and
\emph{v)} that the incoming light only depends on direction, i.e. the illumination is translation-invariant and there are no shadows.
This simplifies the RE to a spherical function
\begin{align}
\label{eq:ReflectanceMap}
L_\mathrm o(\omega_\mathrm o)=
\int_{\Omega^+}
L_\mathrm i(\omega_\mathrm o)
f_\mathrm r(\omega_\mathrm o,\omega_\mathrm i)
\mathrm d \omega_\mathrm i,
\end{align}
which we call the \emph{reflectance map} \cite{Horn1979} of the \emph{illumination} $L_\mathrm i$ and the \emph{surface reflectance} $f_\mathrm r$.
Henceforth, we will simply refer to the surface reflectance model as the \emph{material}. A data-driven BRDF would be an ideal such model, but here it is simplified to the seven-parameter Phong model \cite{Phong1975}
\[
f_\mathrm r(\omega_\mathrm o,\omega_\mathrm i)=
k_\mathrm d
\cdot
\left<
\mathbf n,
\omega_\mathrm o
\right>^+
+
k_\mathrm s
\cdot
\left<
r(\omega_\mathrm i , \mathbf n),
\omega_\mathrm o
\right>^{k_\mathrm g},
\]
where
$k_\mathrm d$ is called the \emph{diffuse color},
$k_\mathrm s$ the \emph{specular color}, and
$k_\mathrm g$ the \emph{glossiness}.

As both the illumination $L_\mathrm i$ and the reflectance map $L_\mathrm o$ are two-dimensional functions of direction, they can be represented as an image using a suitable mapping from the directional to the Cartesian domain, such as the Lambert, latitude-longitude or the mirror-ball mappings~\cite{Debevec1996}.

\mysubsection{Training Data}{TrainingData}
Our training data is a set of synthetic images of spheres with random materials under random illuminations from random views, see \refFig{TrainingData}.

Training materials were taken from the MERL BRDF database \cite{Matusik2003}, in particular the Phong fit made therein.
There are 100 materials overall - in our case 67 were used for training and 33 for testing - including diffuse, glossy, and mirror-like appearances.
\vspace{-0.3cm}
\mycfigure{TrainingData}{
Training data: Every triplet is a sample from the mapping learned.
The observed reflectance map $L_\mathrm o$, a function of the illumination $L_\mathrm i$ and the material $f_\mathrm r$.
}
\vspace{-0.4cm}

As illumination maps we used 70 HDR such maps in total - 60 for training and 10 for testing - from the commercial content supplier HDR Maps~\cite{Dosch2016}.
These images are radiometrically calibrated, \ie differ from the true physical RGB radiance units by only a factor.
We found this not to be the case for other HDR illumination maps found on the Internet.
All illumination maps were re-sampled to 128$\times$128 pixels, which is also the resolution of the maps we will later infer.

View positions are sampled from a random direction in the $xz$-plane with a random declination of $\pm 10^\circ$; an orthographic projection is used.
The shape is always a highly-tessellated sphere with analytic normals.
For rendering we use the full convolution of the illumination map with the Phong model.
This spherical convolution is computationally demanding and to keep it tractable when producing massive training data, it was implemented using GPUs.
The result again is an $128\times 128$ image.
Overall, we produced approximately 50,000 random sample images of synthetically rendered spheres. Note that for the testing set both the material as well as the illumination map are never seen before.

Two variants of the resulting images are kept, with slightly different purposes: an HDR and an LDR variant.
For the HDR variant we first take the logarithm of the RGB data, stored as an 32-bit float image file.
For the LDR variant we simulate the exposure process, as follows:
First we automatically choose an exposure level using the (5,95)-percentiles.
Second, linear radiance values are mapped into the (0,1)-range.
In this range, values are quantized uniformly into 256 values (8-bit).
Finally, the values are mapped back to absolute radiance and stored in a 32 bit float format.
This procedure simulates the information available to a contemporary capturing device with EXIF information (aperture, exposure time, ISO): radiance quantized to 8 bit in an appropriately chosen exposure, allowing to re-scale it to absolute radiance, but with quantization and clipping.

\mysubsection{Network Structure}{NetworkStructure}
Our approach builds on two CNNs: one for illumination and one for material estimation.
An alternative design is to combine both in a joint network. In all networks we used Huber (smooth L1) loss for regression.

\paragraph{Material CNN}
The input to this network is a 2D image of the reflectance map, while the output is a 7-parameter Phong vector.
The design of the network is shown in \refFig{MaterialCNN}.
Overall, the network consists of multiple layers reducing the resolution, followed by several fully-connected layers.
Note that convolutional units are always followed by a ReLU.

\myfigure{MaterialCNN}{
The reflectance network.
Layers are shown as orange boxes with spatial resolution annotated vertically and feature depth horizontally, while convolutional and pooling units are shown as blue circles and fully convolutional units as green hexagons.
The dotted box is the stages shared in a joint material-illumination CNN.
Every convolution is followed by a ReLU, omitted here for simplicity.
}

\paragraph{Illumination CNN}
The input to the illumination network is also a reflectance map, while its output is an illumination map of half the spatial resolution.
The feature spatial resolution is reduced by about one order of magnitude, from 128 to 25, with the middle layers applied in a fully convolutional fashion.
Also, two layers of de-convolution are added, that take intermediate results from the previous same-resolution layers into account.
Doing so, fine spatial details can be preserved.
Again, convolutional units are always followed by a ReLU.

\myfigure{IlluminationCNN}{
The illumination network, following the same visual conventions as  \refFig{MaterialCNN}.
}

\paragraph{Joint Material-Illumination CNN}
Besides the isolated estimation of the illumination and the reflectance (material) as discussed so far, we also experimented with a network estimating both somewhat more jointly.
Such a network shares the first two layers, marked as a dotted box in \refFig{IlluminationCNN} and \refFig{MaterialCNN}.
After this part, the network is split, resulting in two outputs with their independent losses.

\mysubsection{Direct estimation of material under known illumination}{Direct}
While the 2 CNNs explained above can estimate material and illumination separated or jointly, we also investigate an alternative that combines CNNs with classic inverse rendering.
In particular, we explore using the output of the illumination CNN as an input to a classic inverse rendering solution for material estimation.
To this end, we show how a Phong material (\ie reflectance) can be estimated from a reflectance map and known illumination in closed form.

Consider our simplified reflectance map from \refEq{ReflectanceMap}.
As $f_\mathrm r$ is Phong,
\[
L_\mathrm o(\omega_\mathrm o)=
\underbrace{
k_\mathrm d\int
L_\mathrm i(\omega_\mathrm i) \left<\mathbf n,\omega_\mathrm i\right>^+ \mathrm d\omega_\mathrm i
}
_\text{Diffuse}
+
\underbrace{
k_\mathrm s\int
L_\mathrm i(\omega_\mathrm i) \left<(r(\omega_\mathrm i, \mathbf n),\omega_\mathrm o)\right>^{k_\mathrm g}\mathrm d\omega_\mathrm i,
}
_\text{Specular}
\]
a linear combination of a diffuse reflectance map $L_\mathrm d$ and a gloss-dependent specular reflectance map $L_\mathrm s$:
\[
L_\mathrm o(\omega_\mathrm o)=
k_\mathrm d \underbrace{L_\mathrm d(\omega_\mathrm o)}_\text{Diffuse RM}+
k_\mathrm s \underbrace{L_{\mathrm s,k_\mathrm g}(\omega_\mathrm o)}_\text{Specular RM}.
\]
Having observed many pixel samples of $L_\mathrm o$, and having estimated $L_\mathrm i$ using a CNN, $L_\mathrm d$ and $L_{\mathrm s,k_\mathrm g}$ can be computed for all values of $k_\mathrm g$ using a spherical convolution.
Furthermore, if we hold $k_\mathrm g$ fixed, estimating $k_\mathrm d$ and $k_\mathrm s$ is a linear least-squares problem:
Let $\mathbf l_\mathrm o, \mathbf l_\mathrm d, \mathbf l_{\mathrm s,k_\mathrm g}$ be vectors of those pixels for a gloss level $k_\mathrm g$.
So $\mathbf l_\mathrm o=k_\mathrm d \mathbf l_\mathrm d + k_\mathrm s \mathbf l_{\mathrm s,k_\mathrm g}$ or $\mathsf A\mathbf x=b$, where $\mathsf A=(\mathbf l_\mathrm d|\mathbf l_{\mathrm s,k_\mathrm g})$, $\mathbf x=(k_\mathrm s, k_\mathrm d)$, and $\mathbf b=\mathbf l_\mathrm o$.
This can efficiently be solved for $\mathbf x$ for every gloss level $k_\mathrm g$ by inverting a 2$\times$2 matrix.
To finally find the optimal gloss level $k_\mathrm g$, a line search for discrete gloss levels is performed, in our case on 100 levels, logarithmicaly spaced.

This procedure is only workable because the number of non-linear parameters is low in the Phong model and would not scale to more complex material models.
Still, as we show later, estimating Phong parameters and illumination using CNNs is outperforming the estimation of more complex material models.

\mysubsection{Up-sampling}{Upsampling}
Finally, we suggest a post-processing step, where the discontinuity structure of the reflectance map is enforced onto the estimated illumination. This is motivated by the fact that a reflectance map will never show edges which are not also in the illumination \cite{Dror2001}.
At best, it is glossy and will show fewer edges.
Such an up-sampling should leave the edges in the illumination unaffected, if they are not edges in the reflectance map too, but if an edge is found in the reflectance map, the illumination should contain an edge as well.
This is achieved by joint bilateral upsampling \cite{Kopf2007} of the resulting illumination from $64\times 64$ to $128\times 128$ using the reflectance map itself as a guide.
Up-sampling is a post-process orthogonal to the CNN that can increase the visual effectiveness, mainly for very shiny materials.

\mysection{Results}{Results}
In this section, we perform qualitative and quantitative evaluations of the results produced by our approach.
The CNNs used as well as the training and benchmark data will be made publicly available\footnote{Project:~\url{http://homes.esat.kuleuven.be/~sgeorgou/DeLight/}}.

\mysubsection{Quantitative Evaluation}{QuantitativeEvaluation}
First, we evaluate our approach quantitatively, using it in a re-synthesis task and second, on real photographs of reflectance maps.

\paragraph{Re-synthesis benchmark}
Evaluating a successful decomposition is not trivial due to the complex interplay of material (reflectance), illumination, shape and viewpoint.
The most straightforward evaluation considers the $\mathcal L_2$-norm between illumination maps or material (reflectance) parameters.
Regrettably the correlation of parameter differences with visual quality is limited and errors are often over- or underestimated: if the level of specularity is zero, the glossiness becomes irrelevant; appearance is not changing linearly with Phong glossiness, etc.

Instead, motivated by our applications, we suggest an evaluation in terms of the ability to predict novel images under novel views, illumination or material (\refTbl{Main}).
In this protocol, firstly the reflectance map is separated, secondly material or illumination are manipulated, thirdly the view is re-synthesized and finally compared to a reference synthesized from ground truth material and illumination.
Re-synthesis is done the same way as explained in \refSec{TrainingData}.

The manipulations (columns in \refTbl{Main}) are as follows:
\emph{a)} replace the estimated illumination by a point light,
\emph{b)} replace the estimated material by a mirror,
\emph{c)} keep both as-are,
\emph{d)} replace the estimated illumination by a another random natural illumination from a subset of our corpus, not contained in our training data,
\emph{e)} replace the estimated material by a random one from a subset of MERL BRDF \cite{Matusik2003} corpus, not contained in our training data.
The two last approaches result in a distribution of errors, because of which we sample 1000 random variants and then aggregate the errors into an average.

The different approaches (rows in \refTbl{Main}) to separate the reflectance map into illumination and material are as follows:
``{\sc Indep}'' is our approach with independent CNNs for material and illumination.
``{\sc Joint}'' is our joint network.
``{\sc Greedy}'' means to first run our illumination network and second use the optimization procedure explained in \refSec{Direct} to complement it with materials.
``{\sc LN}'' is the method of Lombardi and Nishino \cite{Lombardi2012}, being the state-of-the-art for this task.
A comparison with their work is made, both when using the default values for their priors (``{\sc LN DP}'') and when using no priors (``{\sc LN NP}'') which might depend on the types of illuminations and materials used.

All the above is ran on the HDR data-set (upper part of \refTbl{Main}), as suggested also by Lombardi and Nishino.
Furthermore, we compare to our two independent CNNs, trained on the LDR data-set (last row of \refTbl{Main}).

The final quantitative measure is the image difference between the re-synthesized image produced using our decomposition and the re-synthesized image produced using the ground-truth decomposition. The mean square-error of the logarithm of HDR radiance (MSE) and a colored, multi-resolution structured similarity index \cite{Wang2003} ran on the tone-mapped LDR result (DSSIM) are used to compare the re-synthesized image to the reference.

\begin{table}[ht]
	\centering

	\setlength{\tabcolsep}{1pt}
	\caption{
    Synthetic evaluation of the different approaches \emph{(rows)} for different tasks \emph{(columns)} according to two error metrics \emph{(lower is better)}.
    The images are samples from selected rows and columns.
	The best method for a task is shown in bold.
	}
	\includegraphics*[clip, width = \linewidth]{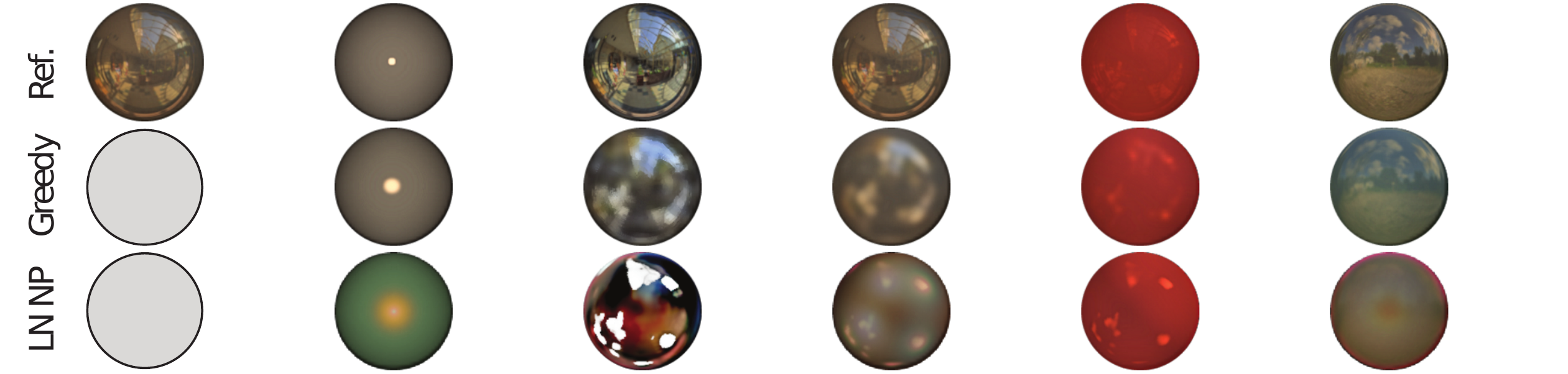}
	\begin{tabular}{l
    m{0.86cm}m{0.97cm}
    m{0.86cm}m{0.97cm}
    m{0.86cm}m{0.97cm}
    m{0.86cm}m{0.97cm}
    m{0.86cm}m{0.97cm}
    }\\
	&
	\multicolumn{2}{c}{Point light\vspace{-0.1cm}}&
	\multicolumn{2}{c}{Mirror Mat.}&
	\multicolumn{2}{c}{Re-synthesis}&
	\multicolumn{2}{c}{MERL Mat.}&
	\multicolumn{2}{c}{Nat.\ Illum.}
	\\
	&
	{\tiny MSE}&{\tiny DSSIM}&
	{\tiny MSE}&{\tiny DSSIM}&
	{\tiny MSE}&{\tiny DSSIM}&
	{\tiny MSE}&{\tiny DSSIM}&
	{\tiny MSE}&{\tiny DSSIM}
	\\
	\toprule
    \emph{HDR}\\
	{\sc Indep.} {\tiny (Our)}
	&{\bf .0055}&.0677&.0603&.1821&.0118&.0685&.0232&.0341&.0006&.0466\\
	{\sc Joint} {\tiny (Our)}
	&.0082&.0753&{\bf .0590}&{\bf .1782}&.0117&.0770&.0200&{\bf .0339}&{\bf .0006}&.0529\\
	{\sc Greedy} {\tiny (Our)}
	&.0062&{\bf .0326}&.0603&.1821&.0016&{\bf .0175}&.0232&.0341&.0008&{\bf .0209}\\ 
	{{\sc LN DP} \cite{Lombardi2012}} 
	&.0245&.1450&.2537&.3299&.0002&.1485&.0288&.0854&.0019&.1423\\
	{{\sc LN NP} \cite{Lombardi2012}} 
	&.0263&.1664&.2862&.3124&{\bf .0001}&.0243&.0292&.0433&.0018&.0605\\
    \midrule
    \emph{LDR}\\
    {\sc Indep.} {\tiny (Our)}
    &.0082&.0691&.0626&.1901&.0011&.0624&{\bf .0027}&.0354&.0006&.0472\\
    \bottomrule
	\end{tabular}
	\label{tbl:Main}
\end{table}

Overall, we find that our methods outperform the state-of-the-art, according to all metrics, with one exception, which is discussed below.

When using our estimated material and re-rendering with a point light, our CNN outperforms competitors by a large factor in MSE and our direct approach by a similar factor according to DSSIM.
Using the estimated illumination and re-rendering on a mirror, is best done using our {\sc Joint} approach, again outperforming competitors by a substantial factor according to both metrics.
According to MSE, re-synthesizing the input image with both the estimated material and illumination, the state-of-the-art approach \cite{Lombardi2012} comes out best.
This is to be expected, as their approach specifically seeks to produce a pair of reflectance and illumination that can be re-synthesized into the input.
According to DSSIM however, which likely is a better measure, our {\sc Joint} approach works best also for this case.
When switching to new materials or illuminations from the corpus the {\sc Joint} network performs best according to all metrics.
The difference to competitors is the strongest in terms of material changes (a three-fold improvement), while re-lighting is only almost twice as good.
Remarkably, the performance of estimating illumination from LDR can be higher than estimating from HDR data, at least according to the MSE measure.
One potential explanation is that the exposure process implies a form of ``whitening", which removes overly bright pixels outside the exposure bracket and consequently helps the convolutions not to over-fit to very large values.
\vspace{-0.4cm}
\mycfigure{Application}{Illumination \emph{(top)} and material change \emph{(bottom)} from estimated RMs \cite{Rematas2016}.}
\vspace{-0.4cm}
\paragraph{Acquired RMs}
The re-synthesis task has been evaluated on the basis of a large choice of variations on a large number of reflectance maps, illuminations and materials.
Capturing this many reflectance maps ourselves is in practice not possible, so we opted for a smaller set of pairs of reflectance and illumination maps where the ground truth illumination is available.
In particular we use a set of 25 materials under 4 different natural illuminations that we have acquired specifically for this task.
They can be found in the supplemental material.

The results are shown in \refTbl{Real}.
The tasks are similar: re-synthesize, but in a more restricted way, as we do not have the ground truth material available; such a task would require a gonioreflectometer.
As the ground-truth illumination is available however (\ie we scanned it using a chrome sphere), we can compute the difference between the ground-truth illumination and the estimated illumination rendered in a mirror (column ``Mirror Mat.'').
Furthermore, we can re-synthesize, using not just a mirror, but instead any material from a database, here again MERL (column ``MERL Mat.'').
Finally, we can predict how a material would look under a different illumination, as the same reflectance maps were captured under this different illumination too, \ie a ``non-parametric re-synthesis'' that is possible as all pairs of materials and illuminations are available.
Note that without ground-truth, re-synthesis under point light illumination is not possible.

For brevity, the state-of-the-art ({\sc LN NP} \cite{Lombardi2012}) is compared to our designs: two independent CNNs ({\sc Indep.}), the joint approach ({\sc Joint.}) and the greedy approach ({\sc Greedy}).

\begin{table}[ht]
	\centering

	\setlength{\tabcolsep}{1pt}

	\caption{Real-data evaluation for the different approaches \emph{(rows)} in terms of three tasks \emph{(columns)} according to two error metrics \emph{(lower is better)}.
    The images are samples from selected rows and columns.
	The best method for a task is shown in bold.
	}
	\includegraphics*[clip, width = \linewidth]{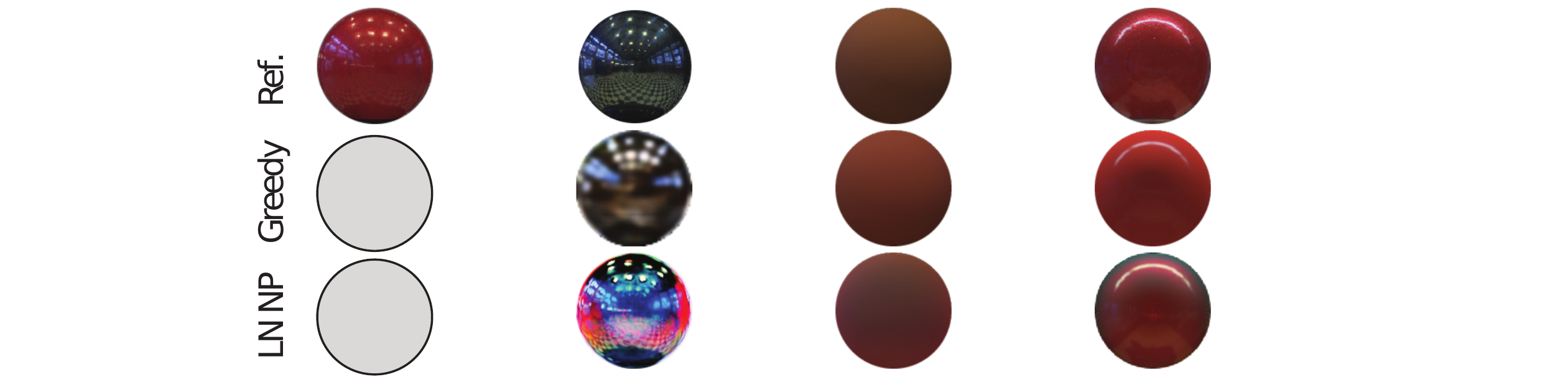}		
	\begin{tabular}{l
    m{1cm}m{1cm}
    m{1cm}m{1cm}
    m{1cm}m{1cm}
    }
	&
	\multicolumn{2}{c}{Mirror Mat.\vspace{-0.1cm}}&
	\multicolumn{2}{c}{MERL Mat.}&
	\multicolumn{2}{c}{Nat.\ Illum.}
	\\
	&
	{\tiny MSE}&{\tiny DSSIM}&
	{\tiny MSE}&{\tiny DSSIM}&
	{\tiny MSE}&{\tiny DSSIM}
	\\
	\toprule
    \emph{HDR}\\   
	{\sc Indep.} {\tiny (Our)} &{\bf 0.929}&0.376&0.099&0.062&1.111&0.183\\
	{\sc Joint} {\tiny (Our)} &0.933&0.376&0.092&0.059&{\bf 1.110}&0.186\\
	{\sc Greedy} {\tiny (Our)} &0.929&0.376&0.099&0.062&1.223&{\bf 0.106}\\
	{\sc LN NP} {\cite{Lombardi2012}}&5.402&0.662&1.722&0.071&3.938&0.187\\
    \midrule
    \emph{LDR}\\   
	{\sc Indep.} {\tiny (Our)} &0.950&{\bf 0.365}&{\bf 0.052}&{\bf 0.043}&1.155&0.214\\
    \bottomrule
	\end{tabular}
	\label{tbl:Real}
\end{table}

We find, that our approach outperforms the state-of-the art by a large margin in terms of MSE and by a good margin for DSSIM.

\mybfigure{Results}{
Visual results:
Please see the text in \refSec{QualitativeEvaluation} for a discussion.
}

\mysubsection{Qualitative Evaluation}{QualitativeEvaluation}
A typical application is interactive reflectance and illumination manipulation as seen in \refFig{Application} and the supplemental video material. Starting from the RM and normal estimation of \cite{Rematas2016}, we re-render the imaged object ($1st$ column) under different illumination ($1st$ row) and different material ($2nd$ row).
Our approach allows for the re-rendering of scenes for different materials \emph{(horizontal variation, \refFig{Results})}, different illuminations \emph{(intra-block vertical variation, \refFig{Results})}, and different shapes \emph{(inter-block vertical variation, \refFig{Results})}.
Decompositions are shown as pairs, where the left half shows re-synthesis using our estimated decomposition, the right column a re-synthesis using reference material and illumination.
The input reflectance maps are marked with a dotted circle.
We see that our approach can reconstruct plausible materials and illumination maps with fine details.

\mysection{Conclusion}{Conclusion}
We have shown how Convolutional Neural Networks (CNNs) can be used to decompose a 2D image of a reflectance map into specular reflectance (material) and complex natural illumination.
This is enabled by training on a large data-set of rendered images, yet it is applicable to synthetic and real images at test time.
It allows for applications where reflectance and illumination are manipulated and used for re-synthesis.
In particular, the generalization capabilities that are important to such manipulation are greatly improved.
Our approach outperforms state-of-the-art baseline in our quantitative and qualitative experiments. 

\bibliographystyle{splncs}
\bibliography{DeepSpecularIntrinsics}
\end{document}

%% file: our-commands.tex
\newcommand{\argmax}[1]{\underset{#1}{\operatorname{arg\,max}}}
\newcommand{\argmin}[1]{\underset{#1}{\operatorname{arg\,min}\ }}
\newcommand{\diag}[1]{\operatorname{diag}(#1)}

\def\figurePath{Figures/}
\def\myfigure#1#2{\begin{figure}[th]\centering\includegraphics*[width = \linewidth]{\figurePath#1}\caption{#2}\label{fig:#1}\end{figure}}
\def\mycfigure#1#2{\begin{figure*}[th]\centering\includegraphics*[clip, width = \linewidth]{\figurePath#1}\caption{#2}\label{fig:#1}\end{figure*}}
\def\mybfigure#1#2{\begin{figure*}[!h]\centering\includegraphics*[clip, width = \linewidth]{\figurePath#1}\caption{#2}\label{fig:#1}\end{figure*}}

\def\mysection#1#2{\section{#1}\label{sec:#2}}
\def\mycsection#1#2{\section*{#1}\label{sec:#2}}
\def\mysubsection#1#2{\subsection{#1}\label{sec:#2}}
\def\mysubsubsection#1#2{\subsubsection{#1}\label{sec:#2}}

\newcommand{\refSec}[1]{Sec.~\ref{sec:#1}}
\newcommand{\refFig}[1]{Fig.~\ref{fig:#1}}
\newcommand{\refEq}[1]{Eq.~\ref{eq:#1}}
\newcommand{\refTbl}[1]{Table~\ref{tbl:#1}}
\newcommand{\unsure}[1]{{\sethlcolor{yellow}\hl{#1}}}

\newcommand{\ie}{i.\,e.\ }
\newcommand{\eg}{e.\,g.\ }
\newcommand{\etal}{\emph{et\,al.}\ }
\newcommand{\citeetal}[1]{et\,al.~\cite{#1}}

\makeatletter
\renewcommand{\paragraph}{%
  \@startsection{paragraph}{4}%
  {\z@}{2.25ex \@plus 1ex \@minus .2ex}{-1em}%
  {\normalfont\normalsize\bfseries}%
}
\makeatother

\newcommand\blfootnote[1]{%
  \begingroup
  \renewcommand\thefootnote{}\footnote{#1}%
  \addtocounter{footnote}{-1}%
  \endgroup
}

%% file: DeepSpecularIntrinsics.bbl
\begin{thebibliography}{10}

\bibitem{Barrow1978}
Barrow, H.G., Tenenbaum, J.M.:
\newblock Recovering intrinsic scene characteristics from images.
\newblock Comp. Vis. Sys. (1978)

\bibitem{Eigen2014a}
Eigen, D., Puhrsch, C., Fergus, R.:
\newblock Depth map prediction from a single image using a multi-scale deep
  network.
\newblock In: NIPS. (2014)

\bibitem{Li2015}
Li, B., Shen, C., Dai, Y., van~den Hengel, A., He, M.:
\newblock Depth and surface normal estimation from monocular images using
  regression on deep features and hierarchical {CRF}s.
\newblock In: CVPR. (2015)

\bibitem{Liu2015}
Liu, F., Shen, C., Lin, G.:
\newblock Deep convolutional neural fields for depth estimation from a single
  image.
\newblock In: CVPR. (2015)

\bibitem{Eigen2015}
Eigen, D., Fergus, R.:
\newblock Predicting depth, surface normals and semantic labels with a common
  multi-scale convolutional architecture.
\newblock In: ICCV. (2015)

\bibitem{Wang2015}
Wang, X., Fouhey, D.F., Gupta, A.:
\newblock Designing deep networks for surface normal estimation.
\newblock In: CVPR. (2015)

\bibitem{Richter2015}
Richter, S., S.Roth:
\newblock Discriminative shape from shading in uncalibrated illumination.
\newblock In: CVPR. (2015)

\bibitem{Rematas2016}
Rematas, K., Ritschel, T., Gavves, E., Fritz, M., Tuytelaars, T.:
\newblock Deep reflectance maps.
\newblock In: CVPR. (2016)

\bibitem{Zbrush}
{Right Hemisphere}:
\newblock {ZBruhs MatCap} (2015)

\bibitem{Dana1999}
Dana, K.J., Van~Ginneken, B., Nayar, S.K., Koenderink, J.J.:
\newblock Reflectance and texture of real-world surfaces.
\newblock ACM Trans, Graph. \textbf{18}(1) (1999)

\bibitem{Debevec1998}
Debevec, P.:
\newblock Rendering synthetic objects into real scenes: Bridging traditional
  and image-based graphics with global illumination and high dynamic range
  photography.
\newblock SIGGRAPH (1998)

\bibitem{Dror2001}
Dror, R.O., Leung, T.K., Adelson, E.H., Willsky, A.S.:
\newblock Statistics of real-world illumination.
\newblock In: CVPR. (2001)

\bibitem{Bell2014}
Bell, S., Bala, K., Snavely, N.:
\newblock Intrinsic images in the wild.
\newblock ACM Trans. Graph. \textbf{33}(4) (2014)  159

\bibitem{Barron2015a}
Barron, J.T., Malik, J.:
\newblock Shape, illumination, and reflectance from shading.
\newblock PAMI (2015)

\bibitem{Lombardi2015}
Lombardi, S., Nishino, K.:
\newblock Reflectance and illumination recovery in the wild.
\newblock PAMI (2015)

\bibitem{Johnson2011}
Johnson, M.K., Adelson, E.H.:
\newblock Shape estimation in natural illumination.
\newblock In: CVPR. (2011)

\bibitem{Horn1979}
Horn, B.K., Sjoberg, R.W.:
\newblock Calculating the reflectance map.
\newblock App. Opt. \textbf{18}(11) (1979)

\bibitem{Haber2009}
Haber, T., Fuchs, C., Bekaer, P., Seidel, H.P., Goesele, M., Lensch, H.P.,
  et~al.:
\newblock Relighting objects from image collections.
\newblock In: CVPR. (2009)

\bibitem{Sloan2001}
Sloan, P.P.J., Martin, W., Gooch, A., Gooch, B.:
\newblock The lit sphere: {A} model for capturing {NPR} shading from art.
\newblock In: Graphics interface. (2001)

\bibitem{RematasCVPR14}
Rematas, K., Ritschel, T., Fritz, M., Tuytelaars, T.:
\newblock Image-based synthesis and re-synthesis of viewpoints guided by {3D}
  models.
\newblock In: CVPR. (2014)

\bibitem{Khan2006}
Khan, E.A., Reinhard, E., Fleming, R.W., B{\"u}lthoff, H.H.:
\newblock Image-based material editing.
\newblock ACM Trans. Graph. \textbf{25}(3) (2006)

\bibitem{zeiler2010deconvolutional}
Zeiler, M.D., Krishnan, D., Taylor, G.W., Fergus, R.:
\newblock Deconvolutional networks.
\newblock In: CVPR. (2010)

\bibitem{lee2009convolutional}
Lee, H., Grosse, R., Ranganath, R., Ng, A.Y.:
\newblock Convolutional deep belief networks for scalable unsupervised learning
  of hierarchical representations.
\newblock In: Proc. ICML. (2009)

\bibitem{Long2015}
Long, J., Shelhamer, E., Darrell, T.:
\newblock Fully convolutional networks for semantic segmentation.
\newblock In: CVPR. (2015)

\bibitem{hypercolumn}
Hariharana, B., Arbelaez, P., Girshick, R., Malik, J.:
\newblock Hypercolumns for object segmentation and fine-grained localization.
\newblock In: CVPR. (2015)

\bibitem{Kulkarni2014}
Kulkarni, T.D., Kohli, P., Tenenbaum, J., Mansinghka, V.:
\newblock Deep convolutional inverse graphics network.
\newblock In: NIPS. (2014)

\bibitem{Tang2012}
Tang, Y., Salakhutdinov, R., Hinton, G.:
\newblock Deep lambertian networks.
\newblock arXiv preprint arXiv:1206.6445 (2012)

\bibitem{Narihira2015b}
Narihira, T., Maire, M., Yu, S.X.:
\newblock Direct intrinsics: Learning albedo-shading decomposition by
  convolutional regression.
\newblock In: ICCV. (2015)

\bibitem{Zhou2015}
Zhou, T., Kr\"ahenb\"uhl, P., Efros, A.:
\newblock Learning data-driven reflectance priors for intrinsic image
  decomposition.
\newblock In: ICCV. (2015)

\bibitem{Narihira2015a}
Narihira, T., Maire, M., Yu, S.X.:
\newblock Learning lightness from human judgement on relative reflectance.
\newblock In: CVPR. (2015)

\bibitem{Debevec1996}
Debevec, P.E., Taylor, C.J., Malik, J.:
\newblock Modeling and rendering architecture from photographs: A hybrid
  geometry-and image-based approach.
\newblock Proc. SIGGRAPH (1996)

\bibitem{Kajiya1986}
Kajiya, J.T.:
\newblock The rendering equation.
\newblock In: ACM Siggraph Computer Graphics. Volume~20. (1986)  143--150

\bibitem{Phong1975}
Phong, B.T.:
\newblock Illumination for computer generated pictures.
\newblock Comm. ACM \textbf{18}(6) (1975)  311--317

\bibitem{Matusik2003}
Matusik, W., Pfister, H., Brand, M., McMillan, L.:
\newblock A data-driven reflectance model.
\newblock ACM Trans. Graph. (2003)

\bibitem{Dosch2016}
HDRMaps.
\newblock https://www.hdrmaps.com/

\bibitem{Kopf2007}
Kopf, J., Cohen, M.F., Lischinski, D., Uyttendaele, M.:
\newblock Joint bilateral upsampling.
\newblock ACM Trans. Graph. (Proc. SIGGRAPH) \textbf{26}(3) (2007)

\bibitem{Lombardi2012}
Lombardi, S., Nishino, K.:
\newblock Reflectance and natural illumination from a single image.
\newblock In: ECCV. (2012)

\bibitem{Wang2003}
Wang, Z., Simoncelli, E.P., Bovik, A.C.:
\newblock Multiscale structural similarity for image quality assessment.
\newblock In: Proc. Sign., Sys. and Comp. Volume~2. (2003)

\end{thebibliography}
